\pdfoutput=1

\documentclass[11pt]{article}

\usepackage{emnlp2021}

\usepackage{times}
\usepackage{latexsym}
\usepackage{multirow}

\usepackage[T1]{fontenc}

\usepackage[utf8]{inputenc}

\usepackage{microtype}

\usepackage{tikz}
\usepackage{tabularx}
\usepackage{amsmath}
\usepackage{amsthm}
\usepackage{amssymb}
\usepackage{paralist}
\usepackage{booktabs}
\usepackage{todonotes}
\usepackage[capitalise,noabbrev]{cleveref}

\usepackage{xspace}

\title{Logical Reasoning with Span-Level Predictions for \\ Interpretable and Robust NLI Models}

\author{Joe Stacey\\
  Imperial College London \\
  \texttt{j.stacey20@imperial.ac.uk}
  \And
  Pasquale Minervini \\
  University of Edinburgh \& UCL\\
  \texttt{p.minervini@ed.ac.uk}
 \AND
  Haim Dubossarsky \\
  Queen Mary University of London \\
 \texttt{h.dubossarsky@qmul.ac.uk} \\
 \And
  Marek Rei \\
  Imperial College London \\
  \texttt{marek.rei@imperial.ac.uk}
 }

\date{}
\begin{document}
\maketitle
\begin{abstract}
Current Natural Language Inference (NLI) models achieve impressive results, sometimes outperforming humans when evaluating on in-distribution test sets. However, as these models are known to learn from annotation artefacts and dataset biases, it is unclear to what extent the models are learning the task of NLI instead of learning from shallow heuristics in their training data.
We address this issue by introducing a logical reasoning framework for NLI, creating highly transparent model decisions that are based on logical rules. Unlike prior work, we show that improved interpretability can be achieved without decreasing the predictive accuracy. We almost fully retain performance on SNLI, while also identifying the exact hypothesis spans that are responsible for each model prediction.
Using the e-SNLI human explanations, we verify that our model makes sensible decisions at a span level, despite not using any span labels during training. We can further improve model performance and span-level decisions by using the e-SNLI explanations during training. Finally, our model is more robust in a reduced data setting. When training with only 1,000 examples, out-of-distribution performance improves on the MNLI matched and mismatched validation sets by 13\% and 16\% relative to the baseline. Training with fewer observations yields further improvements, both in-distribution and out-of-distribution.
\end{abstract}

\section{Introduction}

The task of Natural Language Inference (NLI) involves reasoning across a premise and hypothesis, determining the relationship between the two sentences. Either the hypothesis is implied by the premise (\textit{entailment}), the hypothesis contradicts the premise (\textit{contradiction}), or the hypothesis is neutral to the premise (\textit{neutral}). NLI can be a highly challenging task, requiring lexical, syntactic and logical reasoning, in addition to sometimes requiring real-world knowledge \cite{Dagan_RTE_first_paper}.
While neural NLI models perform well on in-distribution test sets, this does not necessarily mean they have a strong understanding of the underlying task. Instead, NLI models are known to learn from annotation artefacts (or \emph{biases}) in their training data \cite{gururangan-etal-2018-annotation, poliak-etal-2018-hypothesis}. Models can therefore be right for the wrong reasons \cite{mccoy-etal-2019-right}, with no guarantees about the reasons for each prediction, or whether the predictions are based on a genuine understanding of the task. We address this issue with our logical reasoning framework, creating more interpretable NLI models that definitively show the specific logical atoms responsible for each model prediction.
\begin{figure}
    \includegraphics[width=\columnwidth]{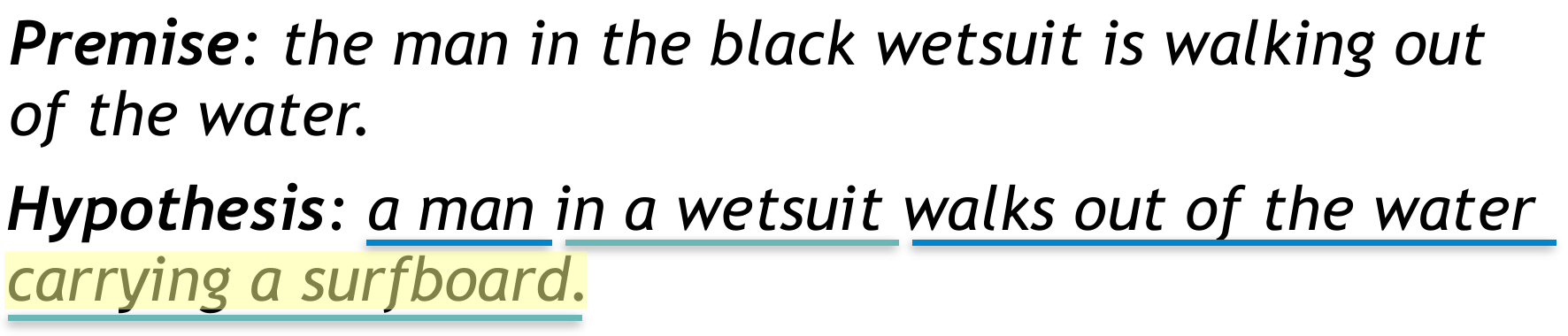} 
    \caption{Example of a hypothesis segmented into spans, with the four different hypothesis spans underlined. The model predicts the neutral class only for `carrying a surfboard.', as this is not implied by the premise.} \label{example_spans}
\end{figure}
One challenge when applying a logical approach to NLI is determining the choice of logical atoms. When constructing examples from knowledge bases, the relationships between entities in the knowledge base can be used as logical atoms \cite{rocktaschel2017endtoend}; in synthetic fact-based datasets, each fact can become a logical atom \cite{clark2020transformers, talmor2020leapofthought}. Neither approach would be suitable for SNLI \cite{bowman-etal-2015-large}, which requires reasoning over gaps in explicitly stated knowledge \cite{clark2020transformers}, with observations covering a range of topics and using different forms of reasoning.
Instead, our novel logical reasoning framework considers spans of the hypothesis as logical atoms. We determine the class of each premise-hypothesis pair based entirely on span-level decisions, identifying exactly which parts of a hypothesis are responsible for each model decision. By providing a level of assurance about the cause of each prediction, we can better understand the reasons for the correct predictions and highlight any mistakes or misconceptions being made. Using the e-SNLI dataset \cite{camburu2018esnli}, we assess the performance of our span-level predictions, ensuring that the decisions about each hypothesis span align with human explanations. 

To summarise our findings:
1) Our Span Logical Reasoning framework (SLR-NLI) produces highly interpretable models, identifying exactly which parts of the hypothesis are responsible for each model prediction. 
2) SLR-NLI almost fully retains performance on SNLI, while also performing well on the SICK dataset \cite{marelli-etal-2014-sick}. This contrasts with previous work, where the inclusion of logical frameworks result in substantially worse performance.
3) Evaluating the SLR-NLI predictions at a span level shows that the span-level decisions are consistent with human explanations, with further improvements if e-SNLI explanations are used during training.
4) SLR-NLI improves model robustness when training in a reduced data setting, improving performance on unseen, out-of-distribution NLI datasets.\footnote{Project code: https://github.com/joestacey/snli\_logic}

\section{Span Logical Reasoning}

Inspired by previous work on error detection \cite{DBLP:conf/naacl/ReiS18, DBLP:conf/aaai/ReiS19, pislar-rei-2020-seeing, kamil}, we construct models for detecting spans of the hypothesis that either contradict the premise (contradiction spans) or are not implied by the premise (neutral spans). We train the model with sentence-level labels while also using auxiliary losses that guide the model behaviour at the span level. As no span labels are provided in the SNLI training data, we supervise our SLR-NLI model at a span level using logical rules, for example requiring a contradiction example to include at least one contradiction span. The model is evaluated based on the span-level decisions for each logical atom.

\subsection{Span-level Approach}

We consider each hypothesis as a set of spans ${s_{1}, s_{2}, s_{3}, ..., s_{m}}$ where each span is a consecutive sequence of words in the hypothesis. For example, in \cref{example_spans} the hypothesis `a man in a wetsuit walks out of the water carrying a surfboard.' contains the following spans: `a man', `in a wetsuit', `walks out of the water',  `carrying a surfboard.'. Each span $s_{i}$ has a label of either entailment, contradiction or the neutral class. In practice, there are no predefined spans in NLI datasets, nor are there labels for any chosen spans. As a result, we propose a method of dividing hypotheses into spans, introducing a semi-supervised method to identify entailment relationships at this span level. In the example provided in \cref{example_spans}, $s_{4}$=`carrying a surfboard.' has a neutral label, while the other spans have an entailment label.

We observe that a hypothesis has a contradiction label if any span present in that hypothesis has a label of contradiction. Similarly, if a hypothesis contains a span with a neutral label and no span with a contradiction label, then the hypothesis belongs to the neutral class. Therefore, a hypothesis only has an entailment class if there are no spans present with span labels of either contradiction or neutral.

When evaluating a hypothesis-premise pair in the test data, our model makes discrete entailment decisions about each span in the hypothesis. The sentence-level label is then assigned based on the presence of any neutral or contradiction spans. This method highlights the exact parts of a hypothesis responsible for each entailment decision. 

\subsection{Span Selection}

We identify spans based on the presence of noun phrases in the hypothesis. Initially, the hypothesis is segmented into spans, with a span provided for each noun phrase which includes both the noun phrase and any preceding text since the last noun phrase. The first span includes any text up to and including the first noun phrase, while the last span includes any text after the last noun phrase. Noun phrases are identified using spaCy\footnote{https://spacy.io}.

However, the most appropriate segmentation of a hypothesis may depend on the corresponding premise, and in some cases, we may need to consider long-range dependencies across the sentence. As a result, we also provide additional spans that are constructed from combinations of consecutive spans. For the example in \cref{example_spans}, this means also including spans such as `a man in a wetsuit' and `walks out of the water carrying a surfboard.'.  We set the number of consecutive spans that are included as a hyper-parameter. We also experiment with a dropout mechanism that randomly masks these additional, consecutive spans for a proportion of the training examples (10\%). This ensures that the model still makes sensible decisions at the most granular span level, while also being able to learn from long-range dependencies across the sentences.

\subsection{Modelling Approach}

A BERT model \cite{devlin-etal-2019-bert} is used for encoding the NLI premise together with each specific hypothesis span, masking the parts of the hypothesis that are not included in the given span. The BERT model provides a [CLS] representation $h_i$ for each span $i$. A linear layer is applied to these representations to provide logits $L_{n,i}$ and $L_{c,i}$ for each span, representing the neutral and contradiction classes respectively.

A separate attention layer is created for both the neutral and contradiction classes that attend to each span-level output. The neutral attention layer attends more to neutral spans, while the contradiction attention layer attends more to contradiction spans. 
Both the neutral and contradiction attention layers consider the same [CLS] representation $h_i$.

The two attention layers use the same architecture, with details provided below for the neutral (\textit{n}) attention layer. Our span-level predictions will be based on the unnormalized attention weights $\widetilde{a}_{n,i}$, which are calculated as:
\begin{equation} \label{eq:attn}
\widetilde{a}_{n,i} = \sigma{(W_{n,2}(\tanh{(W_{n,1} h_{i} + b_{n,1})})+b_{n,2})}
\end{equation}
where $W_{n,1}$, $W_{n,2}$, $b_{n,1}$, and $b_{n,2}$ are trainable parameters. 
\cref{eq:attn} uses a sigmoid so that the output is in the range between 0 and 1 for binary classification.
%
%
Upon normalisation, the attention weights ${a}_{n,i}$ define an attention distribution:
\begin{equation}
a_{n,i} = \frac{\widetilde{a}_{n,i}}{\sum_{k=1}^{m}\widetilde{a}_{n,k}}
\end{equation}
These weights are used to create a logit, $L_{n}$:
\begin{equation} \label{eq:Ln}
L_{n} = \sum_{i=1}^{m}a_{n,i}L_{n,i}
\end{equation}

Using a binary label $y_{n}$, indicating if the example is neutral or not, we create a sentence-level loss to optimise using the sentence labels:
\begin{equation}
\mathcal{L}_{n}^{\text{Sent}} = (\sigma(W_{n,3}\times L_{n} + b_{n,3}) - y_{n})^2
\end{equation}

We combine this with an auxiliary span loss on the model attention weights, $\mathcal{L}_{n}^{\text{Span}}$:
\begin{equation}
\mathcal{L}_{n}^{\text{Span}} = (\max_{i}(\widetilde{a}_{n,i}) - y_{n})^2
\end{equation}

The auxiliary span attention loss has the effect of encouraging the span-level unnormalized attention weights to be closer to zero for entailment examples. This supports our logical framework, which states that all entailment examples must only consist of entailment spans (i.e. with no contradiction or neutral spans). As neutral predictions require at least one neutral span, by supervising the maximum unnormalized attention weight we encourage one of the spans to be predicted as neutral if the sentence label is also neutral. The contradiction-detection attention layer behaves in a similar way, detecting the presence of contradiction spans.

\begin{figure*}[t!]
    \includegraphics[width=500pt]{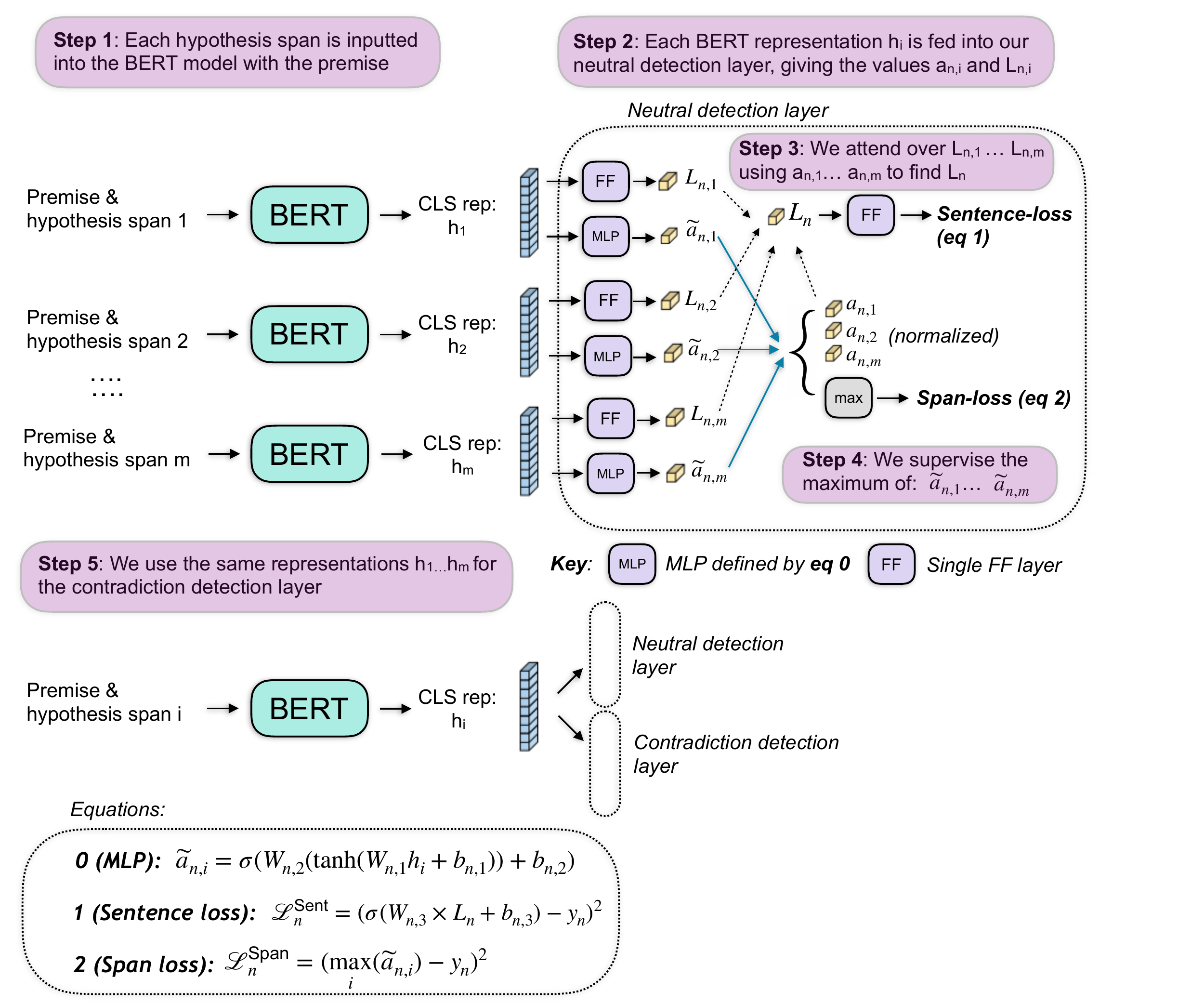}
    \caption{Model diagram describing the training process for SLR-NLI} \label{model_diagram}
\end{figure*}

We then combine together the auxiliary span attention loss with the sentence-level loss:
\begin{equation}
\mathcal{L}_{n}^{\text{Total}} = \mathcal{L}_{n}^{\text{Sent}} + \mathcal{L}_{n}^{\text{Span}}.
\end{equation}
While our model evaluation exclusively makes predictions from the unnormalized attention weights for each span, we find in practice that including a sentence-level objective improves the span-level decisions. In particular, the sentence supervision influences the attention values directly in \cref{eq:Ln}, in addition to supervising the representations $h_i$. The sentence-level supervision does not have access to the full hypothesis, separately considering each span representation $h_i$. See \cref{model_diagram} for a model architecture diagram.

\subsection{Span-level Supervision with Human Explanations}

To provide our model with more information about individual spans, we can use the e-SNLI \cite{camburu2018esnli} human explanations, with rationales highlighting the most important words in each hypothesis. Training models with the e-SNLI explanations can improve both model performance and robustness \cite{joe_and_marek_paper, zhao2020lirex}, although not all prior work has found these improvements \cite{kumar-talukdar-2020-nile, camburu2018esnli, esnli_work_just_after_us}. We assess whether the human explanations can help our model make better decisions at the span level, and also whether the explanations further improve the performance of SLR-NLI.

To incorporate the human explanations during training, we consider the highlighted word rationales for each hypothesis. If any of our SLR-NLI hypothesis spans contain all of the e-SNLI rationale, we assign the overall sentence label as the individual span label. If the hypothesis rationale is not a single consecutive span then we do not provide any supervision with the explanation, as we observe that only single-span rationales consistently align with the desired span-level labels. 

Let $p_{i}$ be the value of 1 where the hypothesis rationale is fully contained within the $i$-th span, and 0 otherwise. Our neutral auxiliary e-SNLI loss $\mathcal{L}_{n}^{\text{e-SNLI}}$ is defined as:
\[\mathcal{L}_{n}^{\text{e-SNLI}} = \lambda^{\text{e-SNLI}}\sum_{i}{p_{i}(\widetilde{a}_{n,i} - y_{n})^2},\]
while $\mathcal{L}_{c}^{\text{e-SNLI}}$ is defined in 
a similar way, using $\widetilde{a}_{c,i}$ and $y_{c}$.

\subsection{Training Process}

Our neutral and contradiction  attention models have two class labels, with $y_{n}$ and $y_{c}$ taking values of 0 or 1. ${y}_{n}=0$ when there are no neutral spans present, while ${y}_{n}=1$ when there is at least one neutral span. ${y}_{c}$ follows the same approach for the contradiction detection label.

For neutral NLI examples, we train our neutral-detection model using a sentence-level label of $y_{n}=1$. Using our logical framework, we also know that a neutral example cannot contain a contradiction span, as any example with a contradiction span would have a contradiction label. Therefore, we train our contradiction-detection model using a sentence-level label of $y_{c}=0$ for these examples. For contradiction examples, we do not train our neutral-detection attention model, as there may or may not be neutral spans present in addition to the contradiction spans. For entailment examples, we train both neutral and contradiction detection models using the labels $y_{n}=0$ and $y_{c}=0$.

Therefore, for neutral or entailment examples we consider the total of both $\mathcal{L}_{n}^{\text{Total}}$ and $\mathcal{L}_{c}^{\text{Total}}$, whereas for the contradiction class we only consider $\mathcal{L}_{c}^{\text{Total}}$.

\subsection{Evaluation}

We evaluate each NLI sentence based exclusively on our span-level decisions. Specifically, an NLI hypothesis is classified as the contradiction class if any of the unnormalized attention weights are predicted as contradiction ($\widetilde{a}_{c,i} > 0.5$).
If there are no contradiction spans present, an NLI example is classified as neutral if there exists at least one neutral span ($\widetilde{a}_{n,i} > 0.5)$. Otherwise, the NLI example is classified as entailment.

\begin{figure}
    \includegraphics[width=\columnwidth]{
    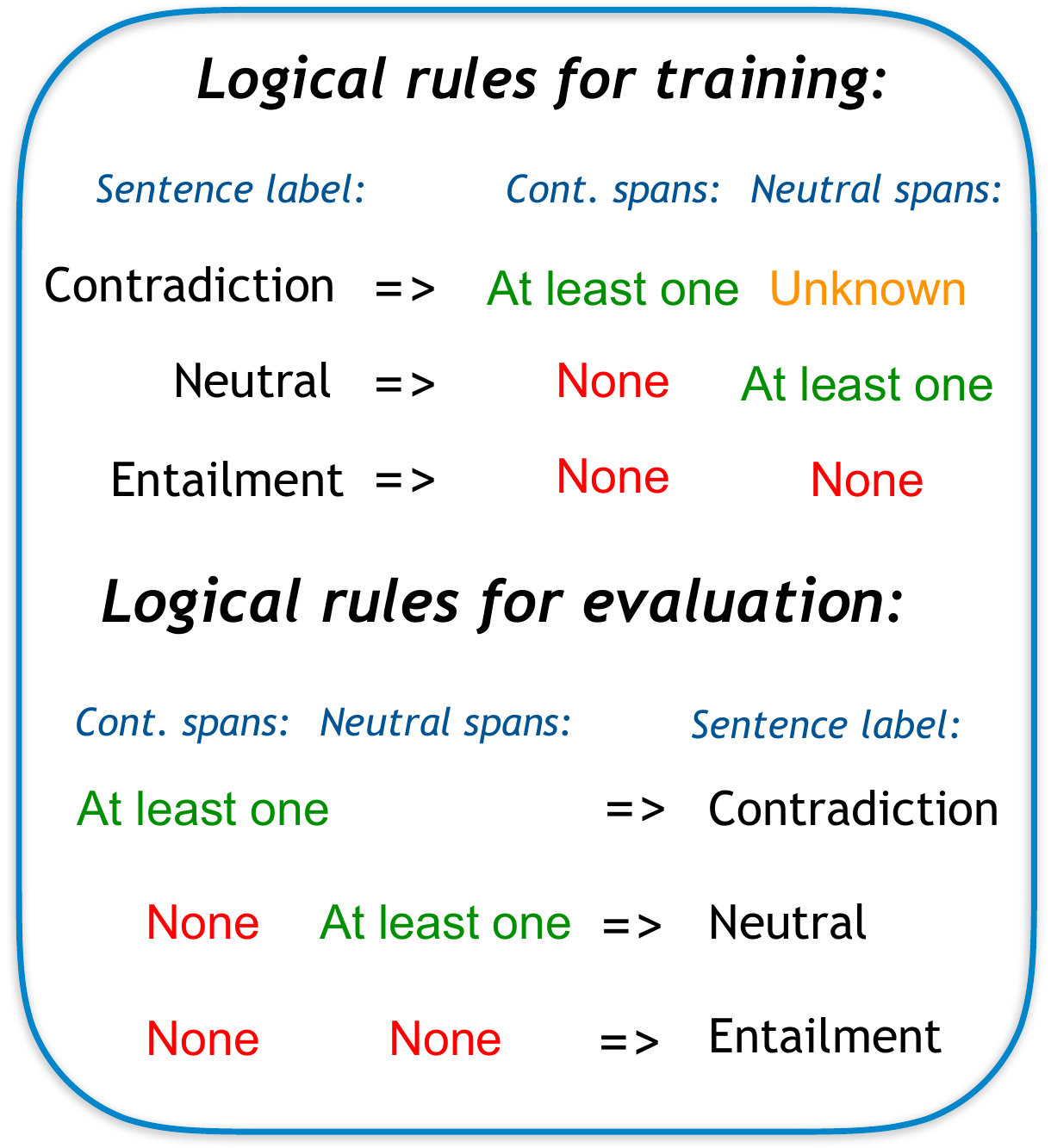} 
    \caption{Our logical rules for training and evaluation. The logical rules for training: 1) for contradiction sentences we use a label of $y_c$ = 1, but do not supervise with a value of $y_n$, 2) for neutral sentences, we supervise with $y_c$ = 0 and $y_n$ = 1, and 3) for entailment sentences we supervise with $y_c$ = 0 and $y_n$ = 0. The logical rules for evaluation: 1) the presence of any contradiction span implies the sentence prediction is contradiction, 2) if no contradiction span is present, the presence of a neutral span implies a neutral sentence prediction, and 3) otherwise we have an entailment prediction.} \label{logic_rules}
\end{figure}
The sentence-level logits $L_{n}$ or $L_{c}$ are only used during training and discarded for evaluation -- they consider information across all the spans and therefore do not allow for a deterministic evaluation of which spans are responsible for the model predictions. The logical rules used in both training and evaluation follow from the inherent nature of NLI (these rules are summarised in \cref{logic_rules}).

\section{Related Work}
\subsection{NLI with Neural Theorem Provers}
Neural theorem provers can effectively solve a range of natural language tasks \cite{rocktaschel2017endtoend,DBLP:conf/acl/WeberMMLR19,minervini2020learning,DBLP:conf/icml/Minervini0SGR20}, many of which could be recast in a similar form to NLI. These datasets are often built from knowledge graphs \cite{sinha2019clutrr, Bouchard2015OnAR, Kok2007StatisticalPI}, for example identifying relationships between characters in short stories \cite{sinha2019clutrr}. Non-neural theorem provers have also shown promising results on the SICK dataset \citep{martinez-gomez-etal-2017-demand, Abzianidze20, Abzianidze17, yanaka-deduction-proofs}, although these methods cannot be easily translated to SNLI, which covers a wide range of topics and uses various forms of reasoning.

\subsection{Monotonic Reasoning with NLI}

Using monotonic reasoning involves matching components of the hypothesis and premise, and using external knowledge from resources including WordNet \cite{WordNet} to 
determine the entailment relationships between corresponding parts of both sentences \cite{Hy-NLI, MonaLog, NeuralLog}. To improve performance, this logical approach can be combined with traditional neural models, learning which examples would benefit from a neural approach rather than using logical rules \cite{Hy-NLI}, or using neural models to decide the class of examples where entailment and contradiction classes have not been detected \cite{MonaLog}. A hybrid approach can improve performance, but at the expense of the interpretability benefits. Logic models using monotonic reasoning are mostly evaluated on SICK and other datasets with a small number of differences between the premise and hypothesis. While our logical framework is not specifically designed for these datasets, we show our performance on SICK still remains competitive with this prior work.

\subsection{Logical Reasoning with SNLI}

Previous work has applied logical reasoning techniques to SNLI, but with performance substantially below baseline levels. \citet{FengRecent} segment a hypothesis into spans, choosing one of seven logical relations for each hypothesis span. A logical relation is predicted for each span using a GPT-2 model \cite{radford2019language} which considers the premise, the given span and all prior hypothesis spans, with reinforcement learning training this span-level behaviour \citep{FengRecent}. Previous work also predicts the seven logical relations for individual words rather than for hypothesis spans \cite{FengOld}.

Closest to our work, \citet{wu2021weakly} label spans as entailment, neutral or contradiction, evaluating at a sentence level based on the presence of neutral or contradiction spans. Our substantial performance improvements compared to \citet{wu2021weakly} reflect our different approaches to supervising at a span level. \citet{wu2021weakly} provide each span model with information about the entire premise and hypothesis, in addition to a hypothesis span and a corresponding premise span. The span label is then predicted using a three-class classifier. 
In comparison, we create separate additional attention layers for neutral and contradiction span detection, combining together multiple different losses to supervise at both the sentence and span level. As we consider neutral and contradiction span detection as separate binary tasks, we also introduce logical rules during training which include not supervising our neutral detection model for contradiction examples, and how a neutral label means there are no contradiction spans present. 

We directly compare our results to \citet{FengOld}, \citet{wu2021weakly} and \citet{FengRecent}.
\section{Experiments}
We train the SLR-NLI model either using SNLI \cite{bowman-etal-2015-large} or SICK \cite{marelli-etal-2014-sick}. 
SNLI is a large corpus of 570k observations, with a diverse range of reasoning strategies required to understand the relationship between the premise and hypothesis. Image captions are used for premises, with annotators asked to create a hypothesis for each class for each given premise. In contrast, SICK has 10k observations and initially uses sentence pairs from image captions and video descriptions, with additional sentence pairs generated by applying a series of rules, including replacing nouns with pronouns and simplifying verb phrases. As a result, entailment and contradiction examples in SICK are often the same except with one or two small changes. Previous work exploits this similarity, using logical reasoning to identify the contradiction and entailment examples \cite{NeuralLog, MonaLog}. Compared to this setting, SNLI provides a more challenging dataset for applying logical reasoning.

We further experiment with training our model in a reduced data setting, motivated by the hypothesis that forcing our model to learn at a span-level will make better use of a smaller number of examples. We expect SLR-NLI to be more robust in a reduced data setting, with existing models known to rely on dataset biases when overfitting to small datasets \cite{utama-etal-2020-towards}. For the reduced data experiments, we train SLR-NLI with 100 and 1,000 examples from SICK or SNLI, evaluating out-of-distribution performance on other unseen NLI datasets including SNLI-hard \cite{gururangan-etal-2018-annotation}, MNLI \cite{williams-etal-2018-broad} and HANS \cite{mccoy-etal-2019-right}. As no explanations are provided for SICK, we only use explanations when training on SNLI (reported as SLR-NLI-eSNLI).

For SICK, when we consider out-of-distribution performance we evaluate on the corrected SICK dataset \cite{MonaLog}, with labels manually corrected by \citet{MonaLog} and \citet{kalouli2017textual}. However, for a fair comparison to previous work, we use the original SICK dataset when evaluating in-distribution performance from SICK.

To validate that the model is making sensible decisions at a span level, we compare the span-level predictions to the e-SNLI human explanations. For each single-span hypothesis rationale in e-SNLI, we consider each model span containing this entire rationale. Each span that does contain the rationale is evaluated, with its span predictions compared to the sentence-level label.

In summary, we consider the following research questions:
1) Does our interpretable SLR-NLI model retain performance on SNLI?
2) Is SLR-NLI a flexible approach that can also work on SICK?
3) Does SLR-NLI improve performance in a reduced data setting?
4) In the reduced data setting, does SLR-NLI also improve robustness?
5) Does SLR-NLI make sensible decisions at a span level?

\section{Results}

\begin{table}[!t]
\begin{center}
\begin{tabular}{rcccccccccc}
\toprule
%
%
{\bf Accuracy} & \textbf{SNLI} & {\bf $\Delta$} \\ 

\midrule
BERT (baseline) & 90.77 & \\
\midrule
\citet{FengOld}  & 81.2 & \textit{-9.57} \\
\citet{wu2021weakly}  & 84.53 & \textit{-6.24} \\
\citet{FengRecent}  & 87.8 & \textit{-2.97} \\
\midrule

SLR-NLI  & 90.33 & \textit{-0.44} \\
\textbf{SLR-NLI+esnli} & \textbf{90.49} & \textbf{\textit{-0.28}} \\

\bottomrule
\end{tabular}

\end{center}
\caption{Performance (accuracy) on the SNLI test-set from SLR-NLI, with and without the additional e-SNLI supervision during training. Each condition is tested across 5 random seeds, including the baseline.}
\label{SNLI_results}
\end{table}
\begin{table}
\begin{center}
\begin{tabular}{rcc}
\toprule
%
%
{\bf Accuracy} & \textbf{SICK} & {\bf $\Delta$} \\ 

\midrule
BERT (baseline) & 85.52 & \\
\midrule
\multicolumn{3}{c}{\bf Hybrid systems} \\
\midrule
\citet{MonaLog}+BERT  & 85.4 & \textit{-0.1} \\
\citet{Hy-NLI}  & 86.5 & \textit{+1.0} \\
\midrule
\multicolumn{3}{c}{\bf Logic-based systems} \\
\midrule
\citet{MonaLog}  & 77.2 & \textit{-8.3} \\
\citet{Abzianidze17}  & 81.4 & \textit{-4.1} \\
\citet{martinez-gomez-etal-2017-demand}  & 83.1 & \textit{-2.4} \\
\citet{yanaka-deduction-proofs}  & 84.3 & \textit{-1.2} \\
\citet{Abzianidze20}  & 84.4 & \textit{-1.1} \\
\citet{NeuralLog}  & \textbf{90.3} & \textbf{\textit{+4.8}} \\
\midrule
SLR-NLI  & 85.43 & \textit{-0.09} \\
\bottomrule
\end{tabular}

\end{center}
\caption{Performance (accuracy) of SLR-NLI on the SICK dataset compared to previous work. Results for SLR-NLI-eSNLI are not provided, as the e-SNLI explanations are specific to SNLI. Results are an average from across 5 random seeds.}
\label{SICK_results}
\end{table}

\begin{table}[!t]
\begin{center}
\begin{tabular}{rcc}
\toprule

\textbf{Dataset} & \textbf{Baseline} & \textbf{SLR-NLI} \\
\midrule
SICK & 81.11 & \textbf{81.33} \\
\midrule
SNLI-dev & 38.50 & \textbf{46.96}$\ddagger$ \\
SNLI-test & 38.17 &\textbf{46.88}$\ddagger$ \\
SNLI-hard & 38.34 & \textbf{44.58}$\ddagger$ \\
MNLI-mismatch. & 40.90 & \textbf{47.85}$\dagger$ \\
MNLI-match. & 39.72 & \textbf{46.51}$\dagger$ \\
HANS & \textbf{53.22} & 50.61 \\
\bottomrule
\end{tabular}

\end{center}
\caption{Accuracy of SLR-NLI compared to a BERT baseline when training with 1,000 SICK examples. The best results are in \textbf{bold}. All results are an average across 5 random seeds. Statistically significant results with $p < 0.05$ are denoted with $\dagger$, while results with $p < 0.01$ are denoted with $\ddagger$, using a two-tailed bootstrapping hypothesis test \cite{efron1993introduction}.}
\label{reduced_data_sick}
\end{table}

\begin{table*}[!t]
\begin{center}
\begin{tabular}{rcccccccc}
\toprule
\multirow{2}{*}{\bf Model} & \multicolumn{2}{c}{\bf In-Distribution} & \multicolumn{5}{c}{\bf Out-of-Distribution} \\
\cmidrule(lr){2-3} \cmidrule(lr){4-8}
  &  SNLI-dev & SNLI-test & SNLI-hard & MNLI-mis. & MNLI-mat. & SICK & HANS \\
\midrule
Baseline & 73.98 & 73.90 & 59.25 & 49.17 & 48.46 & 52.19 & 50.27 \\
\midrule
PoE  & 60.79 & 61.26 & 54.44 & 41.74 & 42.03 & 45.92 & 50.26 \\
Reweight. & 70.69 & 70.86 & \textbf{59.83} & 46.99 & 47.12 & 48.65 & 50.03 \\
Conf Reg. & 57.32 & 57.51 & 49.61 & 38.05 & 38.54 & 38.93 & \textbf{50.84} \\ 
\midrule
SLR-NLI-eSNLI &\textbf{74.22} & \textbf{74.05} & 59.51 & \textbf{57.05}$\dagger$ & \textbf{54.76}$\dagger$ & \textbf{52.23} & 50.00 \\ 
\bottomrule
\end{tabular}

\end{center}
\caption{Accuracy of SLR-NLI-eSNLI compared to a BERT baseline when training with 1,000 SNLI examples. The best results for each dataset are in \textbf{bold}. We compare to a Product of Experts \citep{clark-etal-2019-dont}, Example Reweighting \cite{clark-etal-2019-dont} and Confidence Regularization \cite{conf_reg_paper}. All results are an average across 5 random seeds. For SLR-NLI-eSNLI compared to the baseline, statistically significant results with $p < 0.05$ are denoted with $\dagger$, while results with $p < 0.01$ are denoted with $\ddagger$, using a two-tailed bootstrapping hypothesis test \cite{efron1993introduction}.}
\label{reduced_data}
\end{table*}

\begin{table*}[!t]
\begin{center}
\begin{tabular}{rcccccccccc}
\toprule
%
%
{\bf Model} & {\bf Sent. acc.} & {\bf Span acc.} & {\bf F-macro} & {\bf F-ent} & {\bf F-neut} & {\bf F-cont} \\ 

\midrule
SLR-NLI \textit{(Zero-shot)} & 90.33 & 84.75 & 84.61 & 81.74 & 84.80 & 87.27 \\
SLR-NLI + dropout \textit{(Zero-shot)} & 90.33 & 87.91 & 87.81 & 85.96 & 86.52 & 90.94 \\
\midrule
SLR-NLI-eSNLI \textit{(Supervised)} & \textbf{90.49} & \textbf{88.29} & \textbf{88.17} & \textbf{86.24} & \textbf{86.99} & \textbf{91.28} \\
\bottomrule
\end{tabular}

\end{center}
\caption{Span-level performance of SLR-NLI-eSNLI compared to SLR-NLI. Span performance is evaluated on the three e-SNLI explanations available for each test observation after training on SNLI. A version of SLR-NLI with a dropout mechanism applied is also included. All results are an average across 5 random seeds.}
\label{esnli_performance}
\end{table*}
\subsection{Performance on SNLI and SICK}
SLR-NLI achieves in-distribution results very close to the standard BERT model on the SNLI test set, with 90.33\% accuracy compared to the baseline of 90.77\% (\cref{SNLI_results}).
This result outperforms prior work on logical reasoning for SNLI, as the inclusion of logical frameworks has previously resulted in large drops in performance \cite{FengRecent, wu2021weakly, FengOld}. We achieve this level of performance without training or evaluating on the full premise and hypothesis pairs. When training with the e-SNLI explanations, we see an additional improvement in accuracy (90.49\%).

SLR-NLI compares favourably to prior logical reasoning work on SICK, despite these baselines being specifically designed for this dataset (\cref{SICK_results}). For example, \citet{NeuralLog} aims to bridge the differences between a hypothesis and premise, an approach not possible with SNLI. 
The strong performance of SLR-NLI on both SNLI and SICK shows the flexibility of this approach across different NLI datasets. As SLR-NLI is a model agnostic framework, we also combine SLR-NLI-eSNLI with a better performing DEBERTa model \cite{he2021deberta}. The DeBERTa-base model accuracy is 91.65\%, compared to 91.48\% for SLR-NLI-eSNLI (-0.17\%). This difference in performance is smaller than for BERT (-0.28\%).

\subsection{Reduced Data Setting}
In a reduced data setting, training SLR-NLI-eSNLI with 1,000 SNLI observations, there are significant out-of-distribution improvements on MNLI matched and mismatched with no loss of performance in-distribution (see \cref{reduced_data}). We show that this improved robustness contrasts with common debiasing methods, including Product of Experts \cite{clark-etal-2019-dont, karimi-mahabadi-etal-2020-end}, Example Reweighting \cite{clark-etal-2019-dont}, and Confidence Regularization \cite{conf_reg_paper}, each trained with a hypothesis-only shallow classifier. When training on SICK, we see out-of-distribution improvements on MNLI-matched, MNLI-mismatched, SNLI-dev, SNLI-test and SNLI-hard (see \cref{reduced_data_sick}). As there is no clear hypothesis-only bias for SICK \cite{belinkov-etal-2019-dont}, we do not perform the same comparison to previous robustness work. 
In-distribution improvements are also observed when training with only 100 SNLI observations, where performance is 18\% higher relative to the baseline, with similar results observed when training on SICK. Out-of-distribution improvements also increase when training with fewer observations (complete results are available in the Appendix).

The only dataset where SLR-NLI performed worse than the baseline was HANS, where each hypothesis consists of words that are also in the premise. For example, the premise of `the doctor was paid by the actor' is accompanied by the hypothesis `the doctor paid the actor' \cite{mccoy-etal-2019-right}. In these examples, evaluating on the smaller spans provides no additional benefit.

\section{Analysis}
\subsection{Span-Level Evaluation}
\begin{figure*}[t!]
    \includegraphics[width=380pt]{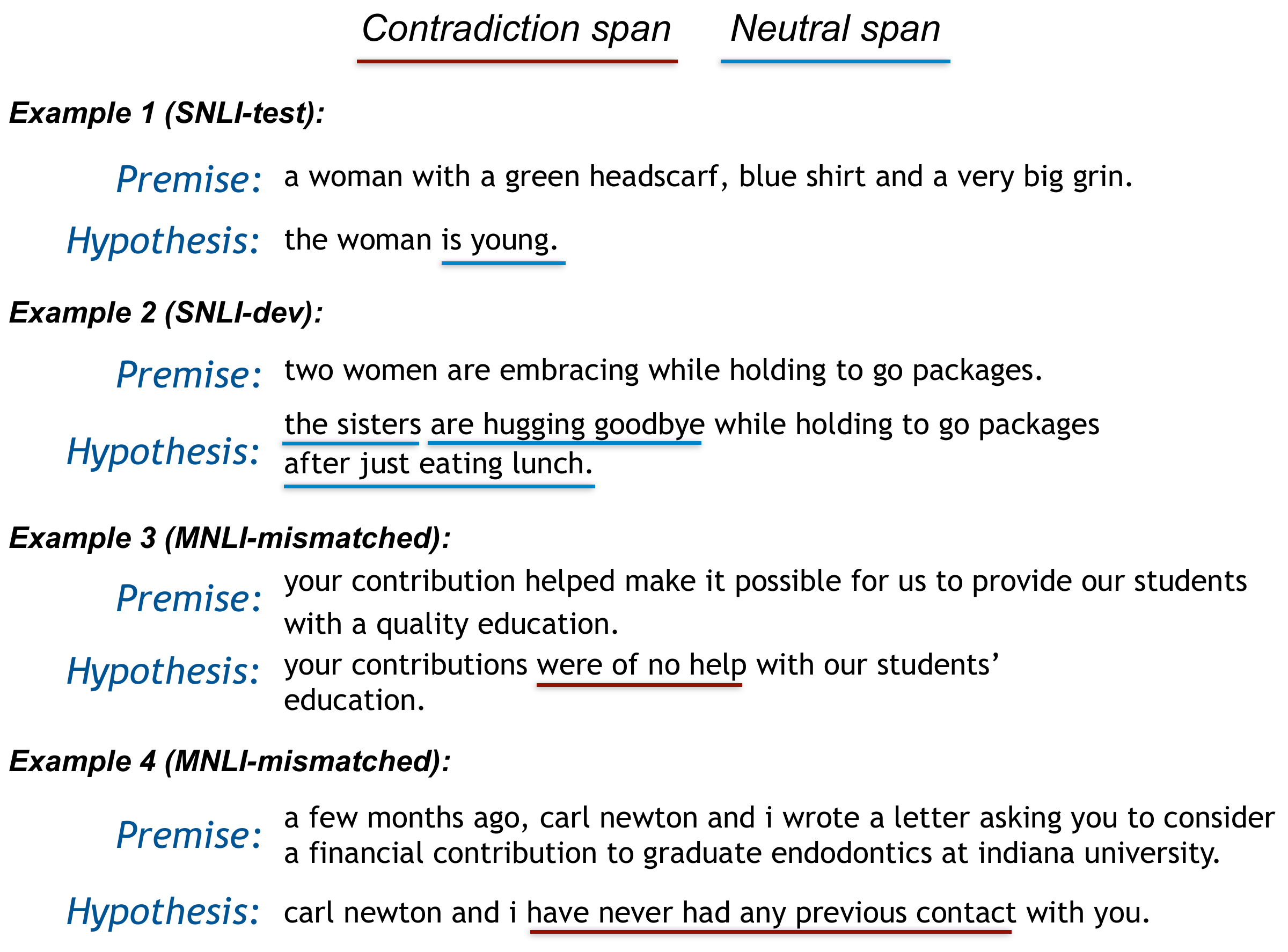} 
    \caption{Span and sentence level predictions for SNLI-test, SNLI-dev and MNLI-mismatched (out-of-distribution). Each of the four examples above are correctly predicted by SLR-NLI-eSNLI.} \label{multiple_example_spans}
\end{figure*}
Even without additional supervision from e-SNLI, SLR-NLI performs well at a span level, with 84.75\% accuracy in a zero-shot setting (\cref{esnli_performance}). This shows that human explanations are not required for the model to make sensible decisions at a span level. With the additional e-SNLI supervision, SLR-NLI-eSNLI reaches a span-level accuracy of 88.29\%. We observe that without the additional e-SNLI supervision, the model tends to rely more on the longer spans that consist of consecutive smaller spans. To mitigate this issue, we experiment with a dropout mechanism during training which randomly masks large spans consisting of consecutive smaller spans, encouraging the model to also make sensible decisions at the most granular span level. In 10\% of observations, all such large spans are masked, leaving only smaller spans as the model input. 
This dropout mechanism improves span performance to 87.91\%, although the sentence-level performance does not improve in tandem (\cref{esnli_performance}).

\subsection{Model Interpretability}

The main advantage of our span-level approach is the interpretability of the model predictions, allowing us to understand which specific parts of the hypothesis are responsible for each predicted label. We define explanation spans as the set of the smallest neutral and contradiction spans such that any longer span that is predicted as neutral or contradiction contains one of these spans. As a result, we only choose longer, multi-segment spans when there are no smaller spans that explain the model decisions. For contradiction predictions, we only include contradiction spans in our explanations.

As shown in \cref{multiple_example_spans}, Example 1, SLR-NLI-eSNLI consistently makes sensible span-level predictions for easier, shorter hypotheses. We therefore show the results of a longer example (Example 2), along with two out-of-distribution examples from the MNLI-mismatched set (Examples 3 and 4). In each case, the model is making correct decisions in line with our human expectations. To provide an unbiased sample of the span-level explanations, we show the first eight neutral and contradiction examples from SNLI-test in the Appendix.

A qualitative analysis shows that some incorrect predictions are a result of subjective labels in the dataset, for example, the model does not find that people walking behind a car implies that the people are necessarily on a street, whereas this example is entailed in SNLI. We also find that the model does not always perform well when evaluating the gender of the people mentioned in the premise. For example, when the premise refers to a girl and `another person', a span of `a man' in the hypothesis can be predicted as contradiction. The model can also predict a span of `a man' to be entailed when a gender is not specified in the premise, for example assuming that a tattooed basketball player mentioned in the premise is a man. This may reveal specific gender biases that are being learnt in the model that would otherwise remain hidden. Finally, the model excels at identifying when multiple different neutral spans are responsible for a neutral classification. This is demonstrated in Example 2 where the three different reasons for neutrality are identified by the model.

\section{Conclusion}
We introduce SLR-NLI as a logical framework that predicts the class of an NLI sentence pair based on span-level predictions. 
SLR-NLI almost fully retains performance on SNLI, outperforming previous logical methods, whilst also performing well on the SICK dataset. The model also outperforms the baseline in a reduced data setting, with substantially improved model robustness. Finally, we create a highly interpretable model whose decisions can easily be understood, highlighting why these predictions are made and the reasons for any misclassifications.

\section{Limitations}

SLR-NLI creates interpretable sentence-level predictions that are based on span-level decisions, however each individual span-level decision is no more interpretable than a standard neural network. We find that this approach provides a balance between maintaining the high performance of neural networks, while also providing explainable NLI decisions.

While SLR-NLI provides a guarantee about which hypothesis spans are responsible for each model prediction, this does not mean that the span-level decisions cannot still be influenced by shallow heuristics. Our approach may help to mitigate some dataset biases, for example the length of the hypothesis \cite{gururangan-etal-2018-annotation} or the proportion of words that overlap between the two sentences \cite{naik-etal-2018-stress}. However, other biases, including how specific words correlate with individual labels  \cite{gururangan-etal-2018-annotation, poliak-etal-2018-hypothesis}, may still be influencing the model. 

Finally, our framework is designed for hypotheses that are mostly a single sentence. Without further modifications, this approach is unlikely to generalise well to more challenging NLI datasets with longer hypotheses, for example with the ANLI dataset \cite{nie-etal-2020-adversarial}.

\section*{Acknowledgements}
We thank our anonymous reviewers for all their thoughtful feedback. We also thank Imperial's LAMA reading group for their support and encouragement, along with Alex Gaskell for his feedback on the project. Pasquale was partially funded by the European Union’s Horizon 2020 research and innovation programme under grant agreement no. 875160, and by an industry grant from Cisco. Haim was partly funded by the research program Change is Key! supported by Riksbankens Jubileumsfond (under reference number M21-0021).

\bibliography{anthology, custom}
\bibliographystyle{acl_natbib}
\appendix

\section{Training in a Reduced Data Setting}

Further experimentation was conducted in a smaller reduced data setting, considering only 100 training examples in SNLI. In this setting we find significant in-distribution improvements for SNLI, with further out-of-distribution improvements on SNLI-hard, MNLI-mismatched, MNLI-matched and SICK (see \cref{reduced_data_100}). With the exception of the SICK dataset, SLR-NLI-eSNLI consistently outperforms the baseline and the three de-biasing methods displayed. Statistical testing was conducted using a two-tailed bootstrapping hypothesis test \cite{efron1993introduction}.

We find similar improvements when testing SLR-NLI on a reduced SICK dataset with only 100 examples. In this case we see better performance for SLR-NLI compared to the baseline for each dataset, with statistically significant improvements in-distribution, in addition to significant improvements on SNLI-dev, SNLI-test and SNLI-hard (see \cref{reduced_data_100_sick}).
\begin{table}[t]
\begin{center}
\begin{tabular}{rcc}
\toprule
 &  \textbf{Baseline} & \textbf{SLR-NLI} \\
\midrule
SICK & 65.31 & \textbf{71.13}$\ddagger$ \\
\midrule
SNLI-dev & 33.58 & \textbf{39.86}$\ddagger$ \\
SNLI-test & 33.41 &\textbf{39.69}$\ddagger$ \\
SNLI-hard & 34.07 & \textbf{39.35}$\ddagger$ \\
MNLI-mismatch. & 35.84 & \textbf{39.99} \\
MNLI-match. & 35.35 & \textbf{39.31} \\
HANS & 49.95 & \textbf{50.73} \\
\bottomrule
\end{tabular}
\end{center}
\caption{Accuracy of SLR-NLI compared to a BERT baseline in a reduced data setting, training with 100 examples from SICK. The best results are in \textbf{bold}. All results are an average across 5 random seeds. Statistically significant results with $p < 0.05$ are denoted with $\dagger$, while results with $p < 0.01$ are denoted with $\ddagger$. This uses a two-tailed bootstrapping hypothesis test \cite{efron1993introduction}.}
\label{reduced_data_100_sick}
\end{table}
\section{Examples of Model Interpretability}

We provide the first eight neutral and contradiction examples within the SNLI test set to show an unbiased sample of the model's span-level explanations (see \cref{appendix_multiple_example_spans}). Entailment examples have not been displayed, as unless these examples have been misclassified, no neutral or contradiction spans would be displayed. With the exception of examples 1 and 8, which are misclassified by the model, the other examples are correct and show span-level decisions in line with our human expectations.

\begin{table*}[ht]
\begin{center}
\begin{tabular}{rcccccccc}
\toprule
 & \multicolumn{2}{c}{\bf In-Distribution} & \multicolumn{5}{c}{\bf Out-of-Distribution} \\
\cmidrule(lr){2-3} \cmidrule(lr){4-8}
  &  SNLI-dev & SNLI-test & SNLI-hard & MNLI-mis. & MNLI-mat. & SICK & HANS \\
\midrule
Baseline & 51.50 & 51.75 & 45.34 & 34.64 & 34.80 & 37.72 & 50.31 \\
\midrule
PoE  & 49.11 & 49.39 & 45.72 & 35.99 & 36.11 & 35.80 & 50.07 \\
Reweight. & 48.66 & 49.01 & 45.56 & 34.95 & 35.28 & 38.44 & 50.04 \\
Conf Reg. & 47.54 & 47.67 & 44.82 & 35.14 & 35.53 & 37.99 & \textbf{50.42} \\ 
\bottomrule
SLR-NLI-eSNLI & \textbf{61.45}$\ddagger$ & \textbf{60.95}$\ddagger$ & \textbf{50.46}$\ddagger$ & \textbf{43.70}$\ddagger$ & \textbf{42.50}$\dagger$ & \textbf{45.77}$\dagger$ & 49.99 \\ 
\bottomrule
\end{tabular}

\end{center}
\caption{Accuracy of SLR-NLI-eSNLI compared to a BERT baseline in a reduced data setting, training with 100 examples. The best results for each dataset are in \textbf{bold}. We compare to a Product of Experts \citep{clark-etal-2019-dont}, Example Reweighting \cite{clark-etal-2019-dont} and Confidence Regularization \cite{conf_reg_paper}. All results are an average across 5 random seeds. For SLR-NLI-eSNLI compared to the baseline, statistically significant results with $p < 0.05$ are denoted with $\dagger$, while results with $p < 0.01$ are denoted with $\ddagger$. This uses a two-tailed bootstrapping hypothesis test \cite{efron1993introduction}.}
\label{reduced_data_100}
\end{table*}

\begin{figure*}[t!]
    \includegraphics[width=380pt]{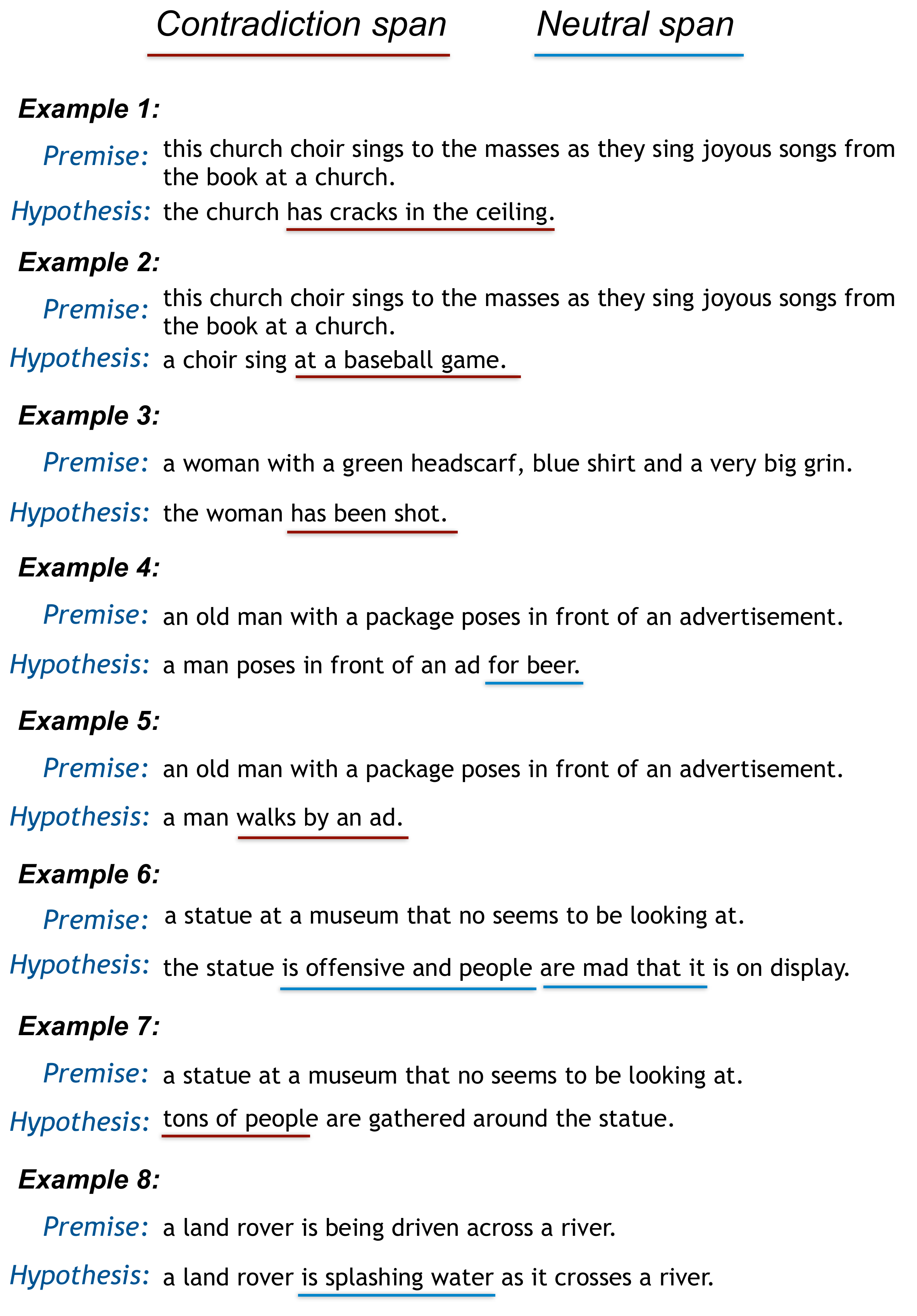}
    \caption{Span and sentence level predictions for the first eight neutral and contradiction examples in SNLI-test. The first example is incorrectly predicted as being contradiction (instead of neutral), while the eighth example is incorrectly predicted as being neutral (instead of entailment). The other predictions are correct and show that the model is making sensible span-level decisions. As the fourth SNLI-test observation is displayed in the main paper we do not repeat this example, and instead we show the ninth example.} \label{appendix_multiple_example_spans}
\end{figure*}

\section{Hyper-Parameter Choices}

We use a BERT-base \cite{devlin-etal-2019-bert} model, providing a direct comparison to previous work. We choose the best learning rate for the baseline, SLR-NLI and SLR-NLI-eSNLI from $\{ 10^{-6}, 2.5\times10^{-6}, 5\times10^{-6}, 7.5\times10^{-6}, 10^{-5} \}$. Each SNLI model is trained over 2 epochs, using a linear learning schedule with a warmup and warmdown period of a single epoch. For the SICK dataset, we train with 3 warmup and 3 warmdown epochs with a learning rate of $10^{-5}$ to reach a baseline comparable with previous work. $\lambda^{\text{e-SNLI}}$ is set as 0.1.
We also consider spans that consist of up to 3 smaller, consecutive spans. A separate hyper-parameter search is conducted for the reduced data setting, with models evaluated with early stopping across 10 epochs. We also perform an additional hyper-parameter search for the DeBERTa-base model \cite{he2021deberta} and for our SLR-NLI-eSNLI model using this baseline. Each hyper-parameter is tested across 5 random seeds, comparing the mean results.

For the baseline BERT model, we find the best performance using a learning rate of $10^{-5}$, whereas SLR-NLI uses a learning rate of $7.5\times10^{-6}$ and SLR-NLI-eSNLI uses a learning rate of $5\times10^{-6}$. For the DeBERTa-baseline, we find the best performance with a learning rate of $7.5\times10^{-6}$, compared to $2.5\times10^{-6}$ for SLR-NLI-eSNLI when using DeBERTa.

For each of our experiments in a reduced data setting, the best performance uses a learning rate of $10^{-5}$ for both SLR-NLI-eSNLI and the baseline. We use the same learning rates when training our model on SICK.

\end{document}